# Improved Exploration in GFlownets via Enhanced Epistemic Neural Networks

Sajan Muhammad*
*Machine Learning Department*
*Mohamed bin Zayed University of Artificial Intelligence*
Abu Dhabi, United Arab Emirates
Corresponding author
mohammadandsaj@gmail.com

Salem Lahlou
*Machine Learning Department*
*Mohamed bin Zayed University of Artificial Intelligence*
Abu Dhabi, United Arab Emirates
salem.lahlou@mbzuai.ac.ae

*Abstract*—Efficiently identifying the right trajectories for training remains an open problem in GFlowNets. To address this, it is essential to prioritize exploration in regions of the state space where the reward distribution has not been sufficiently learned. This calls for uncertainty-driven exploration, in other words, the agent should be aware of what it does not know. This attribute can be measured by joint predictions, which are particularly important for combinatorial and sequential decision problems. In this research, we integrate epistemic neural networks (ENN) with the conventional architecture of GFlowNets to enable more efficient joint predictions and better uncertainty quantification, thereby improving exploration and the identification of optimal trajectories. Our proposed algorithm, ENN-GFN-Enhanced, is compared to the baseline method in GFlownets and evaluated in grid environments and structured sequence generation in various settings, demonstrating both its efficacy and efficiency.

## I. INTRODUCTION

Effectively exploring high-dimensional search spaces remains one of the key challenges in modern machine learning, particularly in scientific discovery applications where the space of possible molecules, materials, or experimental conditions is vast and complex. Estimating *reducible uncertainty* (epistemic uncertainty) has historically played a critical role in enabling efficient exploration in these domains [1], [2], [3], where expensive experiments and simulations make sample efficiency important. Therefore, techniques for quantifying uncertainty and generating diverse, informative sets of experiments are of paramount importance for accelerating scientific discovery. This motivation underlies the development of a relatively new probabilistic machine learning framework called Generative Flow Networks (GFlowNets) [4], [5]. GFlowNets are generative models that sequentially construct objects from a discrete (and more recently [6], [7], continuous) space X by taking a series of actions sampled from a learned policy $P_F$. The policy is trained such that, at convergence, the probability of generating an object $x \in X$ is proportional to a reward $R(x)$ assigned to that object. This property enables GFlowNets to sample diverse, high-reward candidates. This is a crucial capability for scientific applications where we seek not just the single best solution, but a diverse set of promising candidates for further investigation.

GFlowNets are generally trained by sampling trajectories either on-policy (from the learned policy) or off-policy (from a mixture of the learned policy and random noise). Each trajectory terminates at a final object $x \in X$, which is associated with a reward $R(x)$, and a gradient update is performed using this reward signal. Although GFlowNets have shown promise in sampling complex unnormalized distributions and solving combinatorial optimization problems, their training process remains vulnerable to well-known challenges in reinforcement learning (RL), notably *slow temporal credit assignment [8], [9]* and difficulties in balancing *exploration and exploitation*. A key advantage often attributed to GFlowNets is their relative stability under off-policy training, which enables the discovery of multiple modes in the target distribution. This stability stems from their underlying connection to variational inference(VI): recent work has shown formal equivalences between GFlowNet training objectives and hierarchical variational inference, with GFlowNets offering key advantages over traditional VI approaches through their amenability to off-policy training without the high gradient variance typically induced by importance sampling [10]. However, in practice, they frequently become trapped in early-discovered modes, resulting in *mode collapse*, a particularly problematic issue in scientific discovery applications where diverse candidate solutions are essential for comprehensive exploration of chemical, biological, or materials design spaces. This problem is more common in large, sparse-reward environments. This issue highlights the critical need for improved exploration strategies. Without effective exploration, the agent fails to sufficiently interact with the environment, resulting in a suboptimal policy. Since GFlowNets performance is heavily dependent on the quality of the sampled trajectories, devising strategies to encourage more diverse and targeted exploration is essential to achieve better learning outcomes. Despite its importance, the exploration problem in GFlowNets has received comparatively less attention than that given to credit assignment to improve GFlowNet training [11], [12]. As in RL, efficient exploration in GFlowNets involves prioritizing regions of the state space where the learning of the reward distribution is incomplete. However, such exploration must be cost-effective, that is, the benefits of future learning should outweigh the immediate cost of exploratory actions [1], [13]. Achieving this balance requires *uncertainty-driven exploration*, where the agent is guided by an awareness of its own uncertainty about the environment. This uncertainty can be effectively captured by focusing on joint predictions. In many combinatorial and sequential decision making tasks, accurate marginal predictions are insufficient; rather, robust joint predictive distributions are necessary for effective decision-making [15]. To address this, recent work has proposed Epistemic Neural Networks (ENNs) as a principled framework for producing high-quality joint predictions [16]. Empirical results have shown a strong correlation between the quality of joint predictions and improved decision performance. In particular, the *epinet* architecture has been introduced as a way to augment any conventional neural network (including large pretrained models) with minimal computational overhead, while providing calibrated estimates of epistemic uncertainty [16], [17].

In this paper, we explore the crucial role of joint predictions in enabling uncertainty-driven exploration within the GFlowNet framework. To this end, we propose a novel architecture, ENN-GFN, which integrates ENNs into standard GFlowNet training to enhance the quality of joint predictions. Furthermore, we introduce an improved variant, ENN-GFN-Enhanced, designed to further boost exploration efficiency by refining the integration of Thompson sampling inspired techniques as in multi-armed bandits [14]. We evaluate the effectiveness of our proposed methods in two distinct settings. First, we benchmark them against conventional GFlowNet training and a Thompson sampling-inspired approach in the *Hypergrid environment*. Second, we compare the *epinet*-enhanced model with the baseline GFlowNet in a *structured sequence generation task*. In both scenarios, our models consistently outperform the baselines, clearly demonstrating their advantages in tasks that demand efficient and informed exploration.

## II. RELATED WORK

Exploration in RL remains an active area of research, with various strategies being proposed, particularly in uncertainty based exploration methods. A common approach involves using ensembles of multiple independent neural networks, or partially shared architectures, as in [18]. While effective, these methods often come with increased memory and computational costs. Other techniques rely on heuristics, such as those employed in [19], which use the prediction error of a random target function as an exploration bonus. Intrinsically motivated exploration methods also form a prominent class of approaches, one notable example being Random Network Distillation (RND). In this line, [20] proposed augmenting GFlowNets with RND based intrinsic rewards to foster improved exploration. Our work draws inspiration from research that adapts a variant of Thompson Sampling in RL, where the agent maintains a posterior distribution over policies and value functions, and selects actions based on samples drawn from this posterior [13]. Additionally, a key influence on our method comes from the work leveraging approximate Thompson Sampling to generate effective actions by learning reasonably accurate joint predictive distributions using ENNs [17].

## III. PRELIMINARIES AND PROBLEM SETUP

This section introduces the notations and problem formulation for GFlowNets, beginning with a concise overview. A GFlowNet can be represented as a directed acyclic graph $G = (S,A)$, where nodes $s \in S$ denote states and directed edges $(u \to v) \in A$ represent actions. If there exists an edge $(u \to v)$, then $v$ is said to be a child of $u$, and $u$ is a parent of $v$. The graph contains a unique initial state $s_0 \in S$, which has no parent. States with no outgoing edges are considered terminal, and the set of all terminal states is denoted by X. A trajectory $\tau = (s_m \to s_{m+1} \to \cdots \to s_n)$ is a sequence of transitions where each pair $(s_i \to s_{i+1})$ corresponds to an action. A trajectory is said to be complete if it starts at the initial state $s_0$ and ends at a terminal state $s_n \in X$. The collection of all such complete trajectories is denoted by $T$.

The central objective in GFlowNet learning is to discover a forward policy $P_F$ that allows sampling of terminal states with probabilities proportional to a given reward function associated with it. A forward policy is defined as a set of conditional distributions $P_F(\cdot \mid s)$ over the children of every non-terminal state $s \in S$, and it induces a distribution over complete trajectories as

$$P_F(\tau = (s_0 \to \cdots \to s_n)) = \prod_{i=0}^{n-1} P_F\,(si+1 \mid si)$$

.

In addition to the forward policy, a backward policy $P_B$ is introduced, which is a set of distributions $P_B(\cdot \mid s)$ over the parents of every non-initial state $s$. Sampling a terminal state $x \in X$ can be accomplished by starting from $s_0$ and repeatedly sampling actions using $P_F$ until a terminal state is reached. The marginal likelihood of reaching a terminal state $x$ is then given by summing the likelihoods of all complete trajectories that terminate at $x$. Given a reward function $R : X \to R_{\geq 0}$ which is nonnegative and nontrivial, the GFlowNet aims to learn a policy $P_F$ such that:

$$R(x) = Z \sum_{\substack{\tau = T \\ \tau=(s_0\to\cdots\to s_n=x)}} P_F(\tau), \quad \forall x \in X \tag{1}$$

where $Z$ is a normalizing constant equal to $\sum_{x\in X} R(x)$. Since directly computing the summation in Equation (1) is typically infeasible, training procedures involve estimating auxiliary components alongside the forward policy. A prominent training method used is the *Trajectory Balance (TB)* objective [9]. It involves learning not only the forward policy and normalization constant $Z$, but also a backward policy $P_B(s \mid s')$, which approximates the posterior distribution over predecessor states $s$ given a successor state $s'$ within a trajectory. The TB loss for a trajectory $\tau$ is defined as:

$$L_{TB}(\tau;\theta) = \log\left(\frac{Z_\theta \Pi_{t=0}^{n-1} P_F(S_{t+1}|S_t;\theta)}{R(s_n)\Pi_{t=0}^{n-1} P_B(S_t|S_{t+1};\theta)}\right)^2 \quad (2)$$

where $\theta$ represents the parameters of the learned forward policy $P_F$, backward policy $P_B$, and the partition function $Z$. When the trajectory balance loss $L_{TB}(\tau;\theta)$ is minimized to zero for all trajectories $\tau$, the forward policy $P_F$ ensures that terminal states $x \in X$ are sampled in proportion to the reward $R(x)$, thereby satisfying Equation (1) [9], [13].

The goal of this work is to enhance exploration in GFlowNets, thereby improving the efficiency of trajectory selection. Our work is structured to align with this objective. Previous exploration strategies in GFlowNets include On-Policy methods, Tempering, $\varepsilon$-noisy exploration, Generative augmented flow networks [GAFN; 20], and Thompson Sampling-based approaches [13]. Algorithm 1 (Appendix A) is a simplified version of the Thompson sampling-based approach in GFlowNets [13], and is referred to as TS-GFN, to study the importance of uncertainty-aware exploration. TS-GFN illustrates how an agent, equipped with an estimate of its own uncertainty, can more effectively explore less certain regions of the state space, thereby improving chance of sampling trajectories with higher rewards. In this work, we implement three different algorithms to compare their effectiveness in exploration and in selecting appropriate trajectories for training.

### A. TS-GFN

This is a simplified version of the algorithm described in [13]. An approximate posterior over forward policies $P_F$ is maintained by treating the final layer of the policy network as an ensemble. To construct an ensemble of size $K \in Z^+$, the final layer of the policy network is extended to have $K \cdot \ell$ heads, where $\ell$ denotes the maximum number of valid actions for any state $s \in S$. All ensemble members share weights across the layers that precede the final layer. During parameterization of the ensemble of forward policies $K$, a single backward policy $P_B$ is shared between all members of the ensemble $P_{F,k}$. Sharing a common $P_B$ ensures convergence of all $P_{F,k}$ to the same optimal forward policy $P_F$. The remaining procedure is as follows: an ensemble member $P_{F,k}$ is sampled uniformly at random with $k \sim \text{Uniform}\{1,...,K\}$, and a trajectory $\tau \sim P_{F,k}$ is generated. This trajectory is then used to update the selected ensemble member using the trajectory balance loss. The TS-GFN algorithm used in our research is presented in Algorithm 1 (Appendix A).

### B. Joint predictive distributions

Thompson sampling-based work in GFlownets [13] has approached the selection of trajectories for training as an active learning problem, leveraging Bayesian techniques inspired by multi-armed bandits. In particular, TS-GFN maintains an approximate posterior over policies and samples trajectories from this posterior, achieved by interpreting the final layer of the policy network as an ensemble. In decision-making contexts where actions influence future data, the agent's ability to predict outcomes over multiple time steps is critical. Notably, the quality of joint predictions, those that go beyond simple marginal distributions, plays a vital role in striking a balance between exploration and exploitation, as emphasized by [15]. Knowing what they don't know is a key factor in determining the intelligence of an agent. This is best evaluated through the quality of joint predictions [16]. While ensemble methods are capable of providing meaningful joint predictions, their computational demands make them impractical for large models. Conventional GFlowNet algorithms are not inherently suited for efficient joint prediction. Although ensemble-based variants of GFlowNets, offer potential in this direction, they still require evaluation across a broader range of experimental settings [13]. Moreover, ensemble-based approaches come with their own limitations [16]. To better understand Joint predictive distributions an example from [16] is provided in the Appendix F.

### C. ENN and epinet

Given parameters $\theta$ and input $x$, a conventional neural network output $f_\theta(x)$ where $f$ is a parameterized function class. In epistemic neural networks [ENN; 16] the output additionally depends on $z$, which is called an epistemic index sampled from a reference distribution $P_Z$. So an ENN is specified by a parameterized function class $f$ and a reference distribution $P_Z$. Thus, if the output varies with $z$ it indicates the uncertainty which can be resolved with future data, and we can call this *epistemic uncertainty*. Although all standard neural networks can be represented as ENNs, this more general framework allows for richer modeling capabilities, especially in sequential decision-making scenarios [15].

A key strength of ENNs lies in their capacity for joint prediction. In a classification task, for instance, given a sequence of inputs $x_1,...,x_\tau$, a joint prediction assigns a probability $\hat{P}_{1:\tau}(y_{1:\tau})$ to each class combinations $y_1,...,y_\tau$. Using ENNs, one can express such joint predictions via integration over the epistemic index:

$$\hat{P}^{ENN}_{1:\tau}(y_{1:\tau}) = \int_z P_z(dz)\, softmax(f_\theta(x_t, z))\, y_t \quad (3)$$

This approach to joint prediction shares similarities with Bayesian Neural Networks (BNNs), which capture uncertainty by maintaining a posterior distribution over model parameters. Notably, all BNNs can be represented as special cases of ENNs; however, the converse does not necessarily hold and epinet is one such example. The epinet is a special neural network module designed specifically to capture epistemic uncertainty [17]. It is integrated with a base neural network, which is a conventional model parameterized by $\zeta$. Given an input $x$, the base network produces an output $\mu_\zeta(x)$. In addition to the base prediction, the epinet operates on selected features $\phi_\zeta(x)$, which are typically derived from the final layers of the base model, and $z \in R^{DZ}$ called the *epistemic index*, sampled from a standard normal distribution. The parameters of the epinet are denoted by $\eta$, and the overall output of the ENN is expressed as:

$$f_\theta(x,z) = \mu_\zeta(x) + \sigma_\eta(sg[\phi_\zeta(x), z] \quad (4)$$

where $\theta = (\zeta, \eta)$ , and sg denotes the stop-gradient operation applied to the features. The epinet output $\sigma_\eta(\tilde{x}, z)$, where $\tilde{x} := sg[\phi_\zeta(x)]$, is modeled as the sum of two components: $\sigma_\eta(\tilde{x}, z) = \sigma_\eta^L(\tilde{x}, z) + \sigma^P(\tilde{x}, z)$, where $\sigma_\eta^L$ is a *learnable* network, typically a small multilayer perceptron (MLP), that adapts to training data, and $\sigma^P$ is a *fixed prior function* used to encode initial variation across the index $z$, with no trainable parameters. This modular design enables the ENN to express rich epistemic uncertainty while maintaining computational efficiency [16], [17]. The ENN-GFN algorithm used in our research is summarized in Algorithm 2 (Appendix B)

To delve deeper into the fixed prior $\sigma_P$ as explained in [16], the output layer can be interpreted as an ensemble of $D_Z$ networks. Each member of this ensemble is a MLP that takes input $x$ and produces logits for the target classes. Let $p_i(x) \in \mathbb{R}^C$ denote the output of the $i$-th ensemble member. The final output is computed by taking a weighted sum of the ensemble outputs, $\sum_{i=1}^{D_z} pi(x)z_i$, and scaling the result by a tunable factor $\alpha$. This technique is inspired by ensemble-based methods. To further leverage uncertainty awareness and enhance exploration, we introduce a modification of this strategy and refer to the resulting algorithm as ENN-GFN-Enhanced. The algorithm is similar to the strategy explained in [16], with the key difference being that, instead of combining the outputs of the ensemble members using a weighted sum, an alternative strategy is employed: an ensemble member from the prior network is randomly taken and this results in the problem setup similar to maintaining an approximate posterior over forward policies as in Thompson sampling inspired exploration algorithms [13] with a hope that this strategy will exploit uncertainty to promote exploration. Algorithm 3 (Appendix C) summarizes the ENNGFN-Enhanced algorithm employed in our research.

## IV. EXPERIMENTS

### A. Hypergrid Environment

We conduct experiments on the HyperGrid environment as described in [4], [9], [10], [21]. HyperGrid is a $n$-dimensional hypercube grid with side length $H$. The reward function is parametrized with a scalar $R_0$ that determines the difficulty of the exploration problem: a smaller $R_0$ leads to a more challenging target distribution to learn, as it introduces a large region in the state space where rewards are negligible. More details on the environment are provided in Appendix D.

We use torchgfn [21] for our experiments. For comparison purposes, we define a baseline model, referred to as Default-GFN, which corresponds to on-policy TB loss. To ensure a fair comparison, the same loss function and a common base network architecture are used in all algorithms. Specifically, the base network consists of two hidden layers, each with 256 units, and all experiments utilize a uniform backward policy. To evaluate the performance of each method, we measure the $L_1$ distance between the empirical distribution and the target distribution (which can be calculated in closed form in this toy task) over the course of training on the HyperGrid environment.

**2D Hypergrid Environment** : The first experiment is conducted on an 8 × 8 grid with $R_0 = 10^{-4}$. We train the GFlowNet using four different variants: Default-GFN, TS-GFN, ENN-GFN, and ENN-GFN-Enhanced, as described in Section 3. The learned distributions for each algorithm after $10^5$ training trajectories are shown in Figure 2. We illustrate in Figure 1 the evolution of the $L_1$ distance between the target distribution and the empirical distribution. Models trained using ENNGFN-Enhanced and ENN-GFN perform significantly better than the other two variants. More details and discussion are provided in Appendix E1.

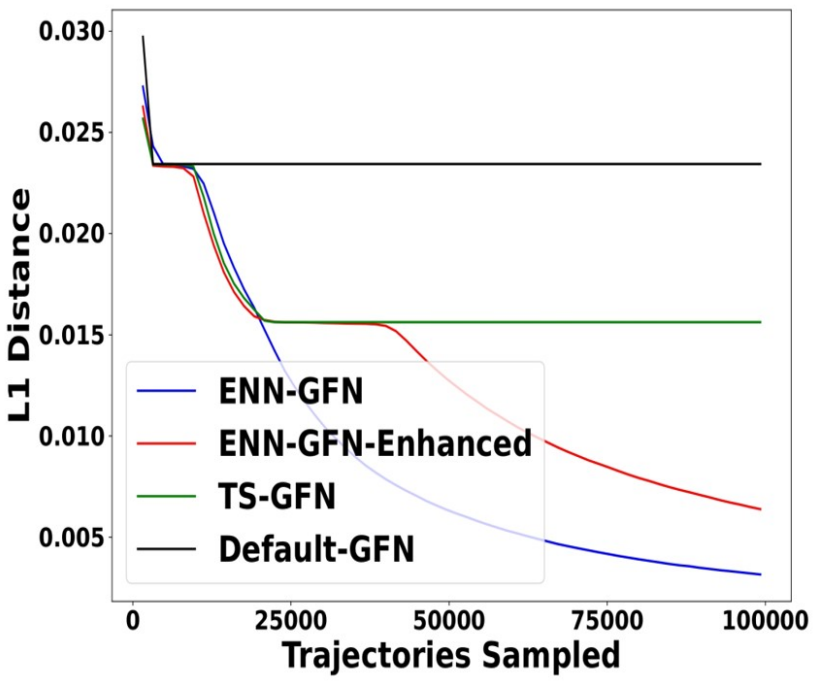

Fig. 1: $L_1$ distance evolution during training.

Sample efficiency: In many interesting settings, querying the reward function is costly. We thus further study the ability of our approach to discover the 4 modes of 8×8 and 16×16 grid environments with $R_0 = 10^{-3}$, using 16,000 and 32,000 sampled trajectories respectively. We find that Default-GFN and TS-GFN struggle to find the 4 modes of the target distributions with such a limited budget, whereas ENN-GFN-Enhanced consistently demonstrates superior performance over the other algorithms, effectively identifying all modes within fewer iterations. The learned distributions and the evolution of the $L_1$ distances are illustrated in Figures 5 and 6 of Appendix E1. Interestingly, ENN-GFN shows relatively poor performance on the 16 × 16 grid within the limited number of iterations and converges to a higher $L_1$, similar to the default GFN. This serves as practical evidence of the enhanced effectiveness of ENN-GFN-Enhanced in more complex environments.

**4D Hypergrid Environment** : The experimental setup follows a configuration similar to the work [4], where a 4D hypergrid environment is used with rewards $R_0 \in \{10^{-1}, 10^{-2}, 10^{-3}\}$, dimensionality $n = 4$, and height $H = 8$. In our experiments, we explore a slightly more challenging setting by using $R_0 = 10^{-3}$ with $H = 16$, and an even more difficult case with $R_0 = 10^{-4}$ and $H = 8$. The results are shown in Figure 3. We find that both ENN-GFN and ENN-GFN-Enhanced learn the target distribution more accurately and faster than other algorithms. We discuss these results in Appendix E2.

**Environment with Increased Complexity :** The environment size plays a critical role in a model's ability to avoid mode collapse. We study this phenomenon in larger 2D HyperGrid environments with highly sparse rewards, setting $R_0 = 10^{-5}$ to increase the likelihood of collapse.

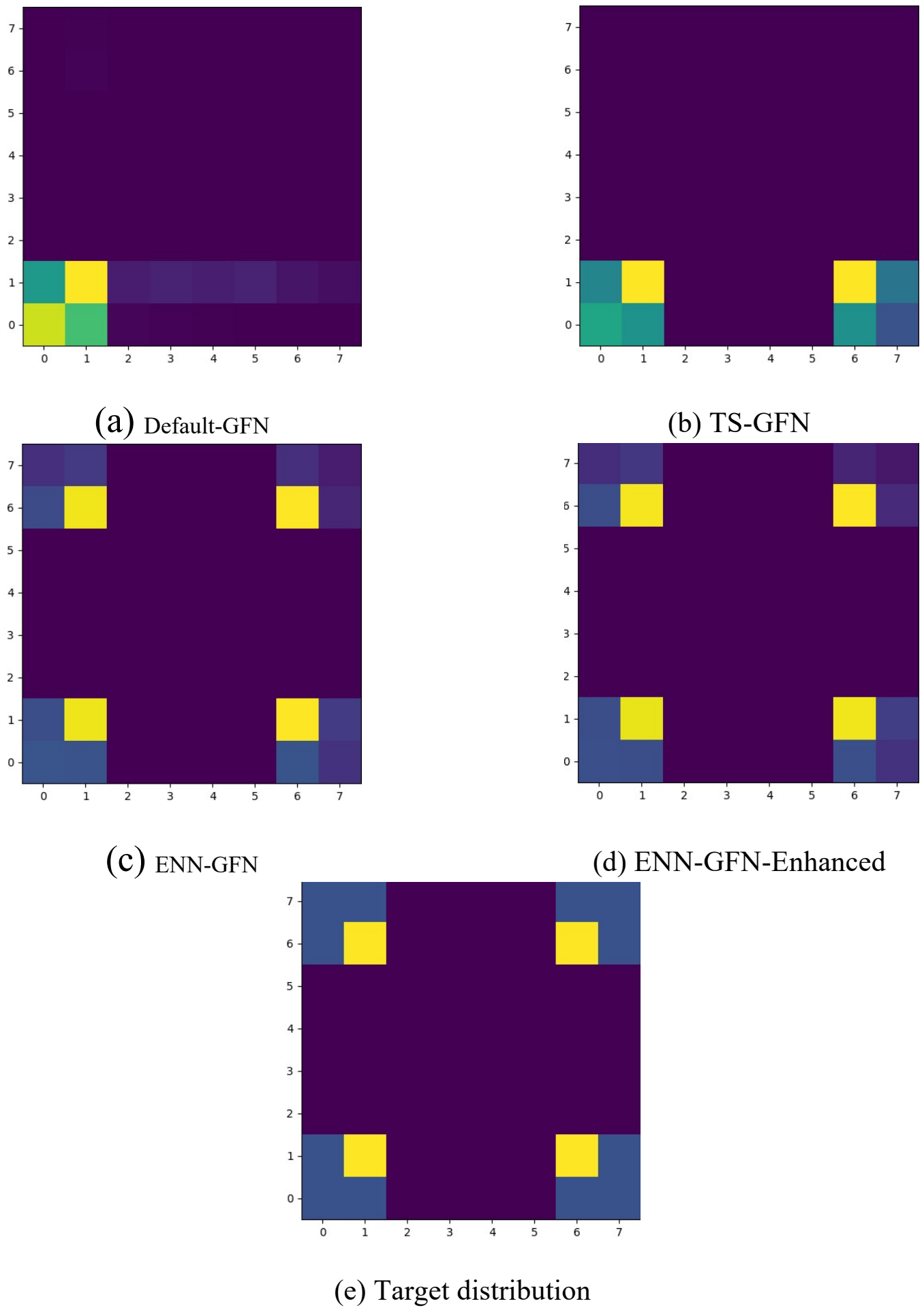

(a) Default-GFN (b) TS-GFN

(c) ENN-GFN (d) ENN-GFN-Enhanced

(e) Target distribution

Fig. 2: Visualization of the learned hypergrid distributions by different methods after sampling $10^5$ trajectories on an $8 \times 8$ grid with $R_0 = 10^{-4}$.

A baseline GFlowNet is trained using the Detailed Balance (DB) objective [5] which we refer to as DB-GFN throughout this work. In the foundational ENN paper, it is noted that the training of specific ENN variants depends on the algorithm developer [16]. This observation motivates our exploration of an alternative loss function. As in prior experiments, all methods use the same loss function and share a common base network architecture to ensure a fair comparison. Specifically, both **ENN-GFN** and **ENN-GFN-Enhanced** are trained using the DB objective, and the ENN parameters $\theta = (\zeta, \eta)$ are updated accordingly, as described in Section 3.3. We discuss the results in Appendix E3.

### *B. Valid bit sequences*

The BITSEQUENCES environment, introduced in [9], serves as a testbed for exploring structured binary sequence generation. In this environment, each state corresponds to a binary sequence, and the agent constructs the sequence by appending either individual bits or blocks of bits.

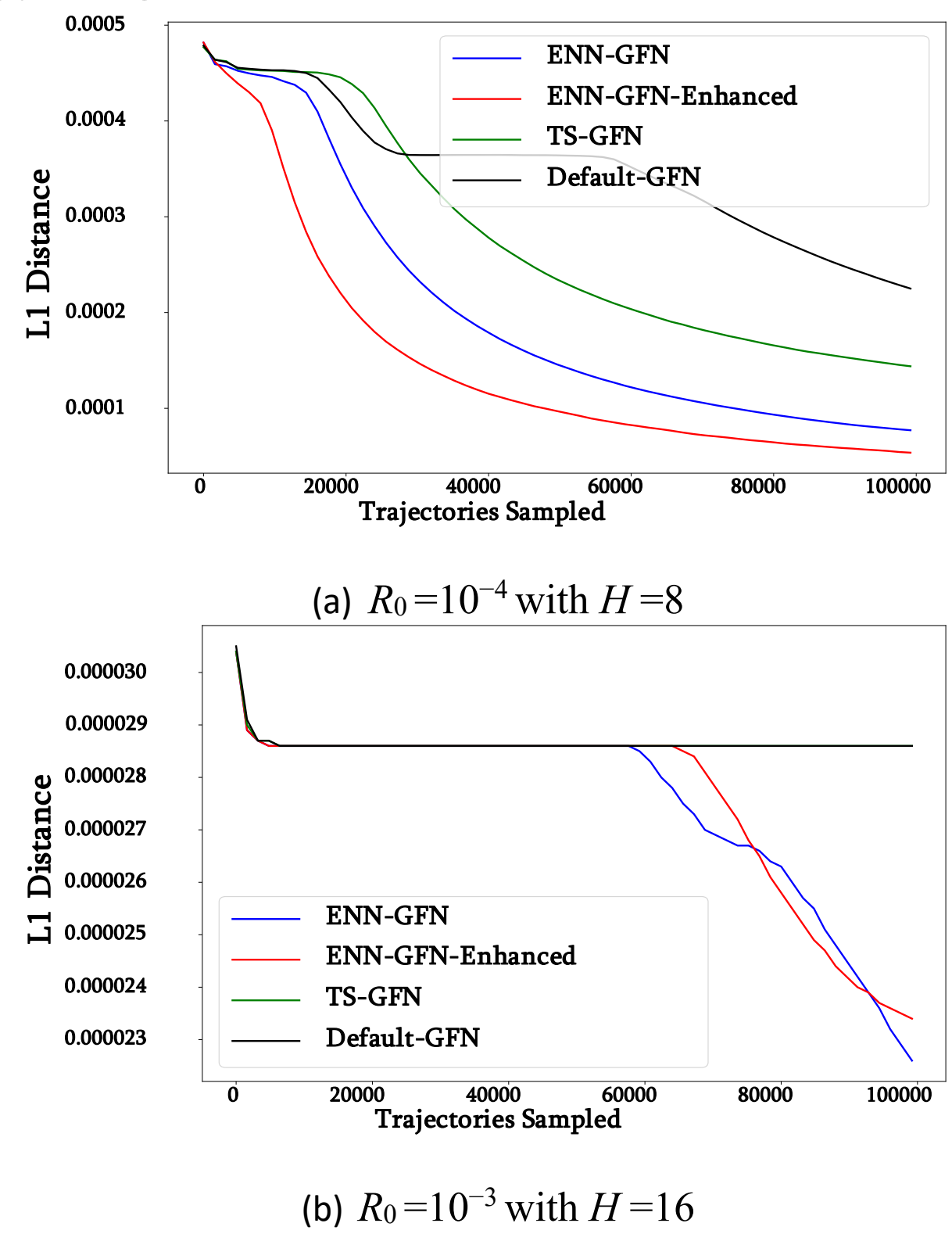

(a) $R_0 = 10^{-4}$ with $H = 8$

(b) $R_0 = 10^{-3}$ with $H = 16$

Fig. 3: $L_1$ distance between the learned and target distributions over the course of training on the 4D hypergrid environment.

The process begins with an empty sequence and terminates once a predefined length is reached. However, in our work, we adopt a different reward structure, as proposed in [22] to address a key limitation identified in the original formulation: namely, that structurally similar sequences not selected as modes are unfairly assigned significantly lower rewards, despite exhibiting the same underlying structure. The new reward structure follows, a valid sequence is defined recursively: it starts with the empty sequence, allows concatenation, and remains valid when a 0 is added to the beginning and a 1 to the end. If 0 is treated as an opening parenthesis and 1 as a closing one, these valid sequences represent balanced parentheses. Sequences reaching the maximum length of 2N are called complete. The goal is to train a GFlowNet to generate all such complete, valid sequences. To assess algorithm performance in this setting, 16,000 sequences are sampled and evaluated using the metrics: diversity. The diversity metric is defined as the fraction of distinct, complete, and valid sequences sampled during generation. For sequences of length 2N, the total number of valid bit sequences corresponds to the Nth Catalan number, which is given by

$$C_N = \frac{1}{N+1}\binom{2N}{N} = \frac{(2N)!}{(N+1)!\,N!}$$

For this task, we use the GPT2Model from Hugging Face's transformers library, configured with 3 hidden layers, 8 attention heads, and an embedding dimension of 64, following the setup as in [22]. We compare the performance of the default Transformer architecture with a modified version that incorporates an epinet. The results, presented in Table 1, demonstrate that integrating epinet into the Transformer significantly enhances its performance.

TABLE I: Comparison of the number of valid balanced parentheses strings generated by transformer-based models with and without epinet integration across varying sequence lengths.

| Model | Len 16 Div ↑ | Len 24 Div ↑ | Len 32 Div ↑ |
|---|---|---|---|
| With Epinet | 260 | 216 | 135 |
| Without Epinet | 207 | 189 | 120 |

## V. CONCLUSION

This paper demonstrates that principled uncertainty quantification through joint predictions significantly improves exploration efficiency in GFlowNets. By integrating Epistemic Neural Networks with standard GFlowNet architectures, our ENN-GFN variants consistently outperform baseline methods across challenging sparse-reward environments, effectively discovering multiple modes and achieving lower $L_1$ distances to target distributions. Our work establishes a foundation for uncertainty-driven exploration in sequential construction tasks, with promising implications for scientific discovery applications where diverse, high-quality candidates are essential. The demonstrated improvements in mode discovery and exploration efficiency suggest this approach could enable more effective search in complex structured spaces, from molecular design to materials discovery.

Future work should explore applications in real-world scientific domains [23], curriculum-augmented applications [24], [25], [26], active learning [27], [28], language modelling [29], [30], and investigate how these uncertainty-aware exploration strategies scale to larger, more complex combinatorial spaces.

## APPENDIX

### *A. TS-GFN Algorithm*

**Algorithm 1** TS-GFN

1: **Input**: Family of $K$ forward policies $\{P_{F,k}\}_{k=1}^{K}$, shared backward policy $P_B$, loss function $L(\tau; P_F, P_B)$
2: **for** each iteration do
3: Initialize batch of initial states from environment
4: Initialize an empty batch of trajectories
5: Sample $k \sim \text{Uniform}\{1,...,K\}$ to select forward policy $P_{F,k}$
6: **while** not in exit state do
7: Use $P_{F,k}$ to sample next actions and states for the batch
8: **end while**
9: Add rewards to the batch of trajectories
10: Compute loss $L(\tau; P_{F,k}, P_B)$
11: Take gradient step on the loss
12: **end for**

### *B. ENN-GFN Algorithms*

Algorithms 2 and 3 show both algorithms we developed.

**Algorithm 2** ENN-GFN

1: **Input**: Output dimension: $d_{\text{out}}$, epinet-train, epinet-prior, base-net, base-head, reference distribution $P_Z$, initialization $\theta_0$, initialize $d_Z$ ensemble layers, forward policy $P_F$, backward policy $P_B$, total number of trajectories $N$, loss function: $L(\tau; P_F, P_B)$
2: **for** $n = 1...N$ do
3: Initialize state $x \leftarrow s_0$, and trajectory $\tau = [s_0]$
4: Sample indices $Z = \{z_1,...,z_{dZ}\} \sim P_Z$
5: **while** $x$ not terminal do
6: Compute shared features: $h \leftarrow$ base-net($x$)
7: Compute base output : base-head(h)
8: Compute: epinet-train-output = epinet-train(h)
9: epinet-train-output= $\Sigma_{j=1}^{dz}\ \text{epinet} - \text{train} - \text{output}_{d_{out},j} . z_j$
10: Compute: epinet-prior-output = epinet-prior(h)
11: epinet-prior-output= $\Sigma_{j=1}^{dz}\ \text{epinet} - \text{prior} - \text{output}_{d_{out},j} . z_j$
12: Compute final output: base output + epinet-train-output + epinet-prior-output
13: $P_F(. \mid x) \leftarrow$ final output
14: Sample next state $x' \sim P_F(. \mid x)$
15: $x \leftarrow x'$ and $\tau \leftarrow [\tau, x]$
16: **end while**
17: Compute loss $L(\tau; P_F, P_B)$
18: Update $\theta$: Take gradient step on the loss
19: **end for**

### *C. ENN-GFN-Enhanced Algorithm*

3 outlines the steps of the ENN-GFN-Enhanced algorithm.

**Algorithm 3** ENN-GFN-Enhanced

1: **Input**: Same as Algorithm 2
2: **for** $n = 1...N$ do
3: Initialize state $x \leftarrow s_0$, and trajectory $\tau = [s_0]$
4: Sample indices $Z = \{ z_1,...,z_{dZ} \} \sim P_Z$
5: **while** $x$ not terminal **do**
6: Compute shared features: $h \leftarrow$ base-net($x$)
7: Compute base output : base-head(h)
8: Compute: epinet-train-output = epinet-train(h)
9: epinet-train-output= $\Sigma_{j=1}^{dz}\ \text{epinet} - \text{train} - \text{output}_{d_{out},j} . z_j$
10: Compute: epinet-prior-output for $d_Z$ ensemble layers
11: epinet-prior-output = output from random ensemble layer
12: Compute final output: base output + epinet-train-output + epinet-prior-output
13: $P_F(. \mid x) \leftarrow$ final output
14: Sample next state $x' \sim P_F(. \mid x)$
15: $x \leftarrow x'$ and $\tau \leftarrow [\tau, x]$
16: **end while**
17: Compute loss $L(\tau; P_F, P_B)$
18: Update $\theta$: Take gradient step on the loss
19: **end for**

### *D. Details on the HyperGrid Environment*

The environment is modeled as a Markov Decision Process (MDP) where the states correspond to the cells of an $n$-dimensional hypercube grid with side length $H$. The agent starts at $\mathbf{x} = (0,0,...,0)$ and is allowed to increment coordinate $i$ by 1 using action $a_i$ (up to a maximum of $H$, at which point the episode terminates). A stop action can also be chosen. Since multiple action sequences can lead to the same final state, the resulting MDP forms a Directed Acyclic Graph (DAG). The reward is defined as:

$$R(x) = R_0 + R_1 \prod_{i=1}^{d} \mathbf{1}\left(0.25 < \left|\frac{x_i}{H} - 0.5\right|\right) + R_2 \prod_{i=1}^{d} \mathbf{1}\left(0.3 < \left|\frac{x_i}{H} - 0.5\right|\right) < 0.4,$$

where $0 < R_0 < R_1 < R_2$. In our experiments, we set $R_1 = 1$ and $R_2 = 3$. By varying $R_0$ and setting it closer to zero, the exploration problem becomes more challenging, as it introduces a large region of the state space where rewards are negligible, discouraging exploration [12, 10, 11].

### *E. Additional Experimental Details and Discussion*

In all experiments, we use a batch size of 16. All other hyperparameters follow the defaults of the torchgfn library.

*1)* ***2D Hypergrid Environment****:* In the $8 \times 8$ setting with $R_0 = 10^{-4}$ and 100,000 training trajectories, ENN-GFN-Enhanced and ENN-GFN perform significantly better than the other variants. Even with limited trajectories, both algorithms capture all modes in the four corners. In contrast, Default-GFN tends to discover only modes near the starting corner.

For the simpler $R_0 = 10^{-3}$ setting with constrained budget, Figure 5 shows the learned distributions for 8×8 and 16×16. Figure 6 shows the evolution of the $L_1$ distance between empirical and target distributions.

*2) **4D Hypergrid environment**:* For the case of $R_0 = 10^{-3}$ and $H = 16$, the default GFlowNet algorithm struggles to discover more number of modes, resulting in the identification of only a subset of them and a correspondingly high $L_1$ distance. Although TS-GFN improves over the default, both ENN-GFN and ENN-GFN-Enhanced perform significantly better in terms of identifying more modes. In the more challenging scenario with $R_0 = 10^{-4}$ and $H = 8$, the task becomes artificially harder as $R_0$ approaches zero. Nevertheless, both ENN-based algorithms are able to recover all modes more effectively and achieve lower $L_1$ distances compared to the other methods.

*3) **Environment with Increased Complexity:*** The experiment is conducted on a 64 × 64 and 128 × 128 HyperGrid environment with a sparse reward setting of $R_0 = 10^{-5}$. We train GFlowNets using three variants of the Detailed Balance objective: DB-GFN, ENN-GFN, and ENN-GFN-Enhanced. Due to the large size and extreme sparsity of the environment, exploration becomes particularly challenging, increasing the likelihood of mode collapse. Nonetheless, ENN-GFN-Enhanced is able to approximate the target distribution better than DBGFN, as shown in Table II.

TABLE II: $L_1$ distance comparison across different algorithms on 2-Dimensional Grid environments (values in $\times 10^{-5}$). $L_1$ distances are obtained after sampling $2 \times 10^5$ trajectories

| Size | DB-GFN | ENN-GFN | ENN-GFN-Enhanced |
|---|---|---|---|
| 64 | 40.00 | 38.00 | 36.00 |
| 128 | 9.74 | 9.14 | 9.13 |

## F. Joint predictive distributions

To understand the difference between marginal and joint prediction, consider the following example from [16]. Suppose a conventional neural network is trained to predict whether a randomly selected person would classify a given drawing as a 'rabbit' or a 'duck'. Given a single image, the network produces a marginal prediction, assigning probabilities to each class. If it outputs 0.5 for both 'rabbit' and 'duck', it is unclear whether this uncertainty arises from genuine perceptual ambiguity across individuals (i.e., some people see a rabbit while others see a duck), or if the model simply lacks sufficient data to learn a definitive classification. Conventional neural networks do not distinguish between these two cases. However, joint predictions—which model the distribution over pairs of labels (e.g., $(y_1, y_2)$) for the same image—can provide this insight. From any such joint distribution, Bayes' rule allows us to compute the conditional distribution, which reveals how one label influences another. By analyzing this conditional, we can assess whether additional training data would reduce uncertainty, or whether ambiguity would persist.

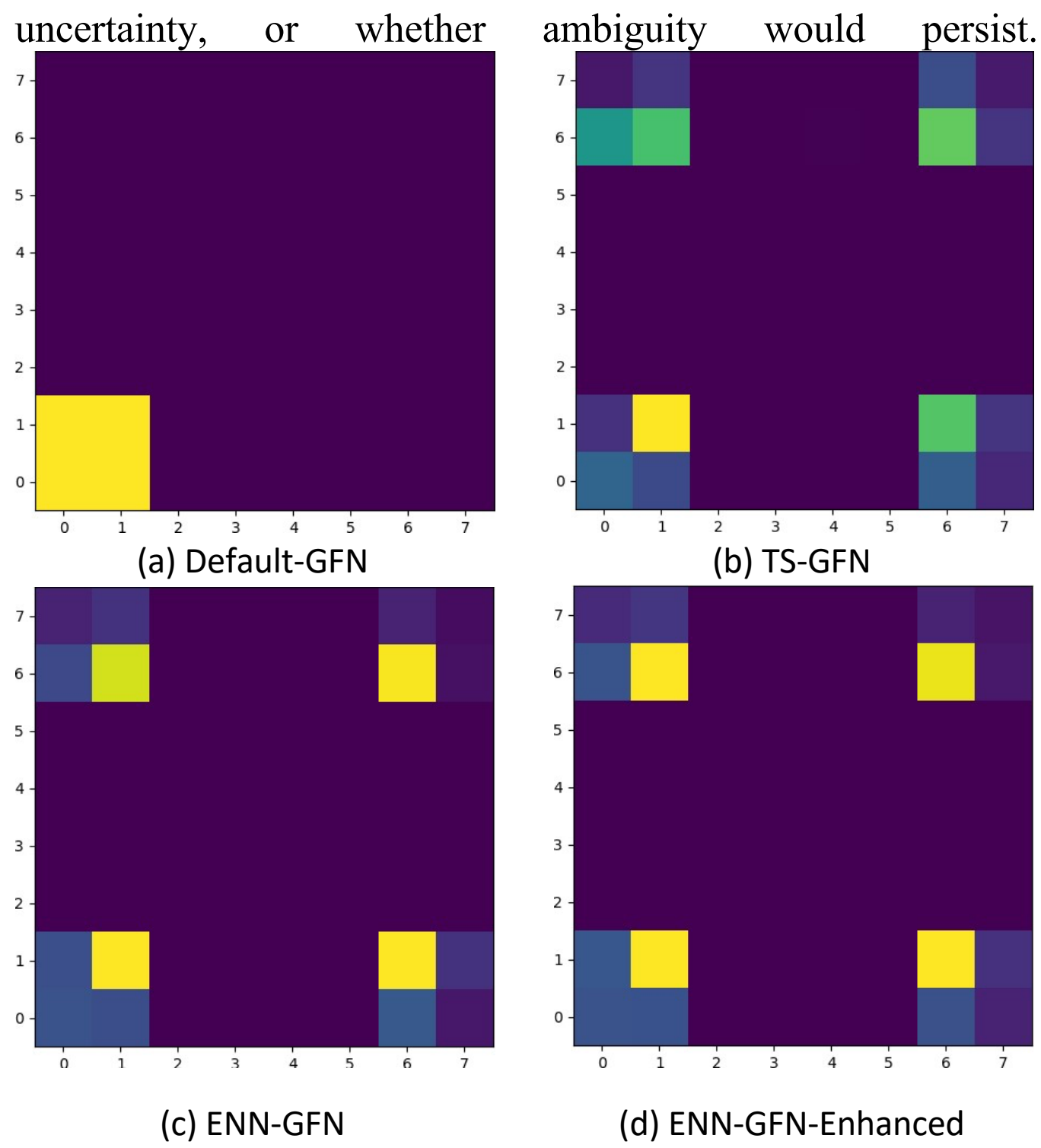

Fig. 5: Visualization of the learned hypergrid distributions by different algorithms, illustrating how quickly they approximate the true distribution within 16,000 iterations for grid size of 8×8 (top row).

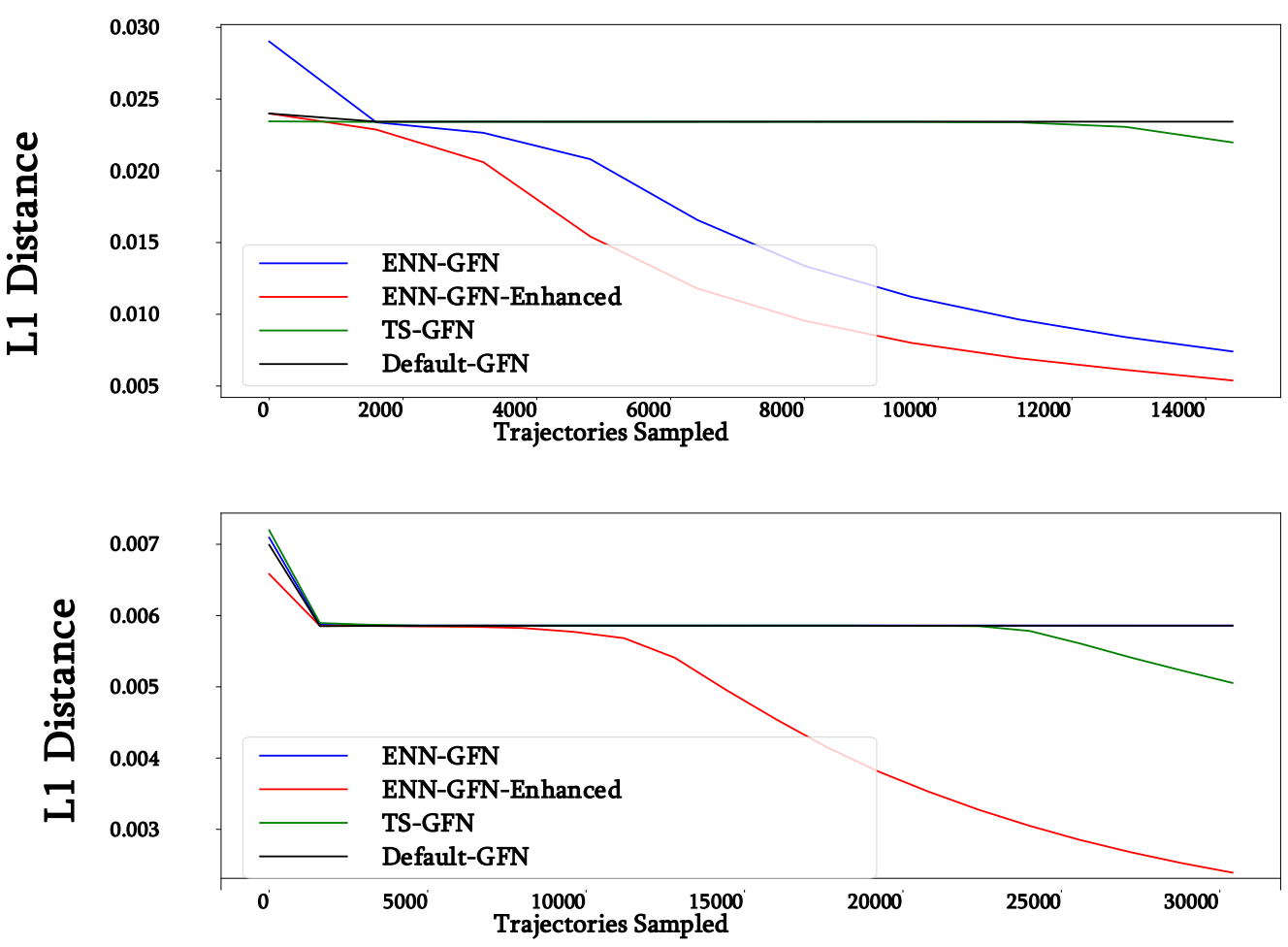

Fig. 6: $L_1$ distance between the learned and true distributions over the course of training on the given hypergrid environments. Left: Performance of the algorithms within 16000 iterations on the 8 × 8 grid task, with $R_0 = 10^{-3}$. Right: Performance of the algorithms within 32000 iterations on the 16×16 grid task, with $R_0 = 10^{-3}$. Across both cases, ENN-GFN-Enhanced consistently demonstrates superior performance.